\RequirePackage[T1]{fontenc}
\documentclass[letterpaper,10pt,conference]{ieeeconf}
\IEEEoverridecommandlockouts
\usepackage{cite}
\usepackage{amsmath,amssymb,amsfonts}
\usepackage{graphicx}
\usepackage{caption}
\usepackage{booktabs}
\usepackage{multirow}
\usepackage{textcomp}
\usepackage{xcolor}
\usepackage[hidelinks]{hyperref}
\usepackage{balance}

\graphicspath{{figures/}}
\hypersetup{
  pdfauthor={Feng Tian, Yulin An, Chenhao Li,
             Matthias Heyrman, Marco Hutter}
}

\newcommand{\teaserfigure}{%
  \vspace{-2em}
  \centering
  \includegraphics[width=1.0\textwidth]{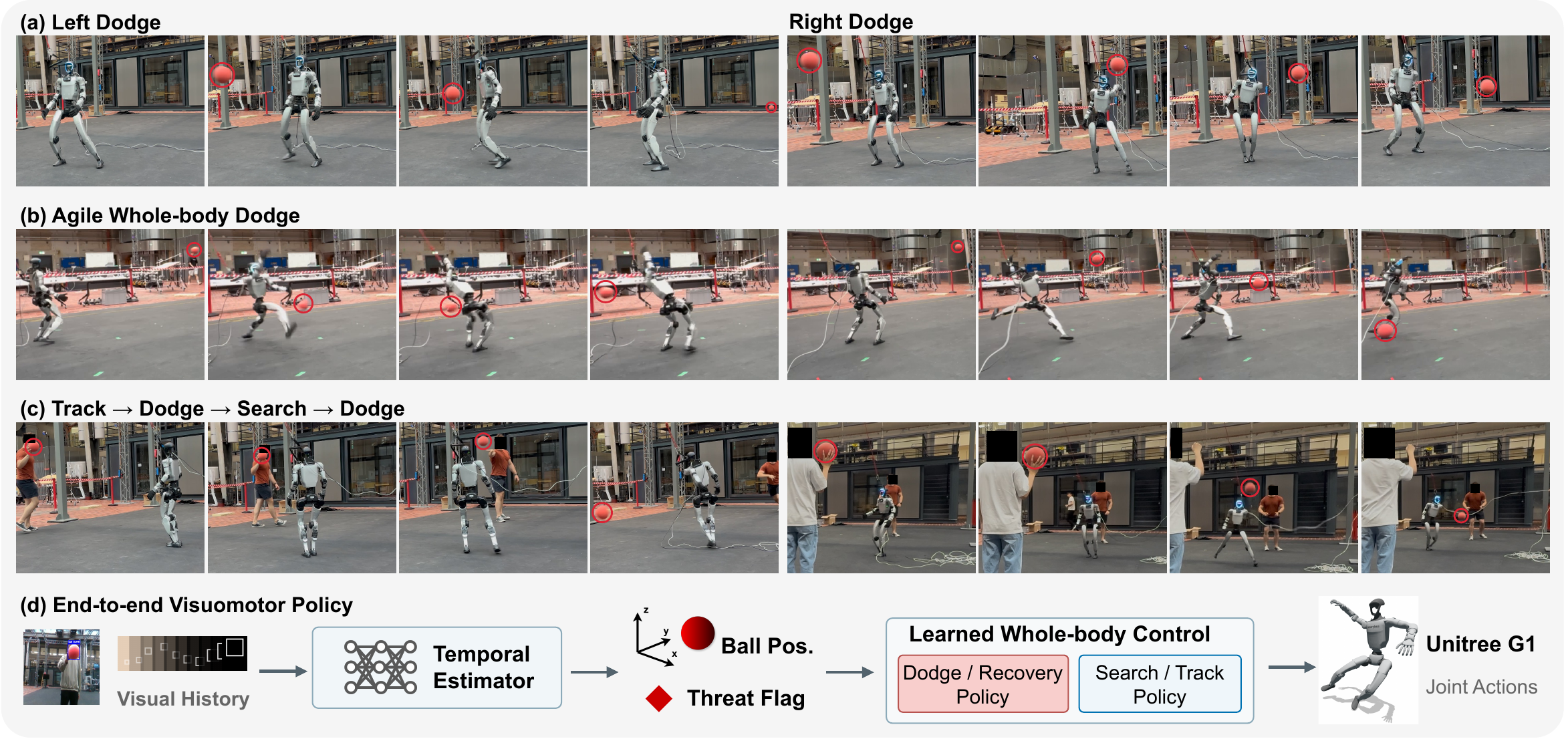}
  \captionof{figure}{Our end-to-end visuomotor system combines temporal visual encoding and threat-flag prediction with learned whole-body skills to enable agile evasion and a repeated search--track--dodge loop on a physical Unitree G1.}
  \label{fig:hardware} 
}

\let\oldtwocolumn\twocolumn
\renewcommand{\twocolumn}[1][]{%
  \oldtwocolumn[{#1\begin{center}\teaserfigure\end{center}}]%
}

\begin{document}

\title{\LARGE\bfseries Learning Agile Humanoid Dodgeball with Onboard Monocular Vision}

\author{%
Feng Tian$^{1,*}$,
Yulin An$^{1,2,*}$,
Chenhao Li$^{3}$,
Matthias Heyrman$^{3}$,
Marco Hutter$^{3}$%
\thanks{%
$^{*}$\,Equal contribution.
$^{1}$\,ETH Zurich, Switzerland.
$^{2}$\,RWTH Aachen University, Germany.
$^{3}$\,Robotic Systems Lab, ETH Zurich, Switzerland.%
}%
}

\maketitle

\begin{abstract}
Humanoid robots operating in shared, dynamic environments must respond to rapidly approaching objects while maintaining balance. We study dodgeball as a controlled test of this capability, coupling visual threat assessment with whole-body evasion, recovery, and target reacquisition. We develop a system that integrates onboard monocular perception, temporal ball-state and threat-flag estimation, and learned whole-body control. A ball detector supplies bounding-box histories, and a temporal estimator predicts relative ball position and a binary threat flag. Separate dodge/recovery and search/track policies share this interface and are composed to support repeated encounters. A three-stage curriculum first trains control with privileged ball states, then starts estimator supervision from stable policy rollouts, and finally replaces the actor's privileged inputs with estimator outputs. Analytic projection and perception randomization provide detector-like training observations without image rendering. Across approach-angle bins in controlled simulation, dodge success ranges from 97.8--100.0\% for 5.0~m launches at 6~m/s to 33.2--40.8\% for 2.5~m launches at 10~m/s, with 1000 throws per speed--distance condition. On a physical Unitree G1, controlled single-throw success exceeds 85\% over more than 15 trials. A hardware sequence demonstrates full-azimuth target reacquisition followed by another dodge. These results demonstrate an integrated onboard-monocular-vision system under highly dynamic motions.

\end{abstract}

\section{INTRODUCTION}
Humanoid robots are intended to operate in environments where people and objects move unpredictably. An accidentally thrown or falling object can enter the robot's immediate workspace before a nominal task controller has time to respond. Avoiding such an object requires the robot to alter its whole-body motion while preserving balance, then recover a posture from which it can continue operating. For a biped, an evasive action that avoids contact but causes a fall is not a successful response. Fast-object avoidance therefore provides a concrete setting for studying reactive whole-body behavior under simultaneous sensing and balance constraints.

We study humanoid dodgeball as a controlled experimental task for this capability. Ball launches make approach speed, distance, and direction explicit, enabling repeatable tests of the time available for perception and movement. The task extends beyond an isolated dodge: the robot must find the ball, assess its approach, evade it, recover, and reacquire a target for the next encounter. A torso-mounted camera makes these requirements interdependent. Turning the body changes visibility, but rapid turning can also degrade the images needed to guide subsequent motion. The tested ball and launch conditions define the scope of our evaluation; they do not establish avoidance of arbitrary projectiles.

Fast-object avoidance has been demonstrated on aerial robots with frame and event cameras~\cite{kaufmann2018deep,wang2021autonomous,falanga2020dynamic,bhattacharya2024monocular}, and quadruped systems combine learned evasion with active sensing~\cite{xu2025rebot,xu2025aprebot}. Humanoid ball-interaction systems demonstrate agile whole-body skills~\cite{ren2025goalkeeper,liu2025humanoidbadminton,su2025hitter,hu2025pace,ren2026smash,zhang2026latent}. Closely related, PAC-MAN combines perception-aware control-barrier-function guidance and reinforcement learning for humanoid dodgeball~\cite{yang2026pacman}. These advances motivate examining the complete sensing-to-action loop, including recovery and target reacquisition with a moving onboard camera.

Our system integrates three components: monocular ball perception, temporal ball-state and threat-flag estimation, and whole-body policy learning. At deployment, an RGB detector supplies ball bounding boxes; a temporal estimator converts their history into relative position and a binary threat flag; and separate dodge/recovery and search/track policies execute the required whole-body behaviors. Their composition supports a search--track--dodge sequence without external ball tracking. The control policies also receive proprioception, so visual task information is coupled to the robot's physical state.

The training design addresses the coupling between these components. Analytic sphere projection generates detector-like observations in parallel simulation, and observation randomization exposes the estimator and policies to noise, missed detections, and delay. A three-stage curriculum separates control acquisition from estimator learning before closing the perception--control loop: the actor first learns from ground-truth ball states, estimator supervision begins after stable rollouts emerge, and the actor then switches to estimated states. The bounding-box interface supports this transition while matching the output available from the deployed detector.

The main contributions are:
\begin{itemize}
    \item We combine analytic 2D bounding-box projection and perception randomization with temporal estimation of relative ball position and a binary threat flag.
    \item We develop a humanoid dodgeball system using onboard monocular vision, separate dodge/recovery and search/track policies, and a three-stage curriculum.
    \item We validate perception, estimation, and avoidance in simulation and on a physical Unitree G1 across controlled launch speeds, distances, and directions.
\end{itemize}

\begin{figure*}[t]
    \centering
    \includegraphics[width=0.82\textwidth]{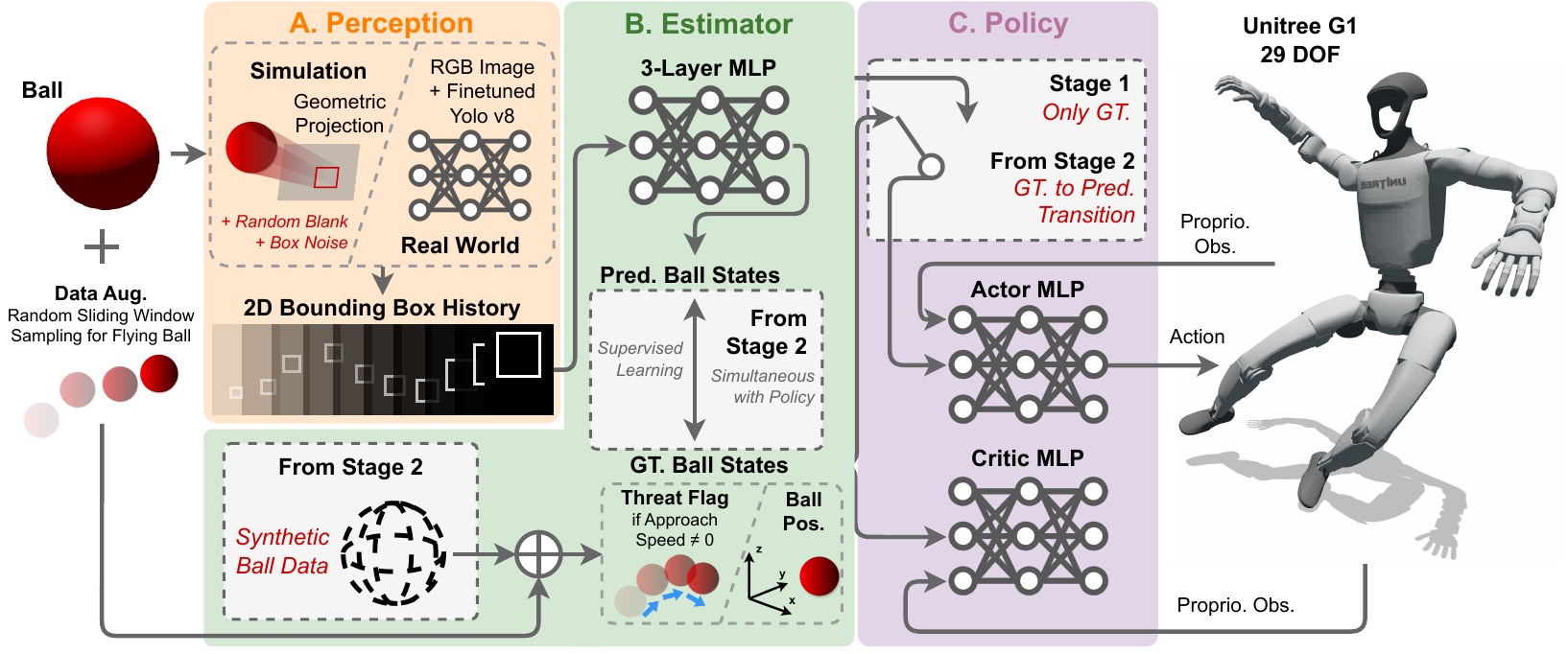}
    \caption{Perception-in-the-loop system. Simulation uses analytic projection and perception randomization to produce bounding-box histories; hardware uses onboard RGB and a fine-tuned detector. The temporal estimator predicts relative ball position and a binary threat flag. The policy block denotes the active whole-body policy: separate dodge/recovery and search/track policies share the estimated-state, proprioceptive, and action interfaces. Ground-truth (GT) states drive the actor in Stages 1--2 and supervise the estimator from Stage 2; the actor uses estimator outputs from Stage 3.}
    \label{fig:overview}
    \vspace{-12pt}
\end{figure*}

\section{RELATED WORK}
\subsection{Dynamic Obstacle Avoidance}
Agile aerial navigation systems learn vision-based control or combine perception, prediction, and kinodynamic planning~\cite{kaufmann2018deep,wang2021autonomous}. Event cameras are attractive for projectiles because their high temporal resolution reduces latency and motion blur~\cite{falanga2020dynamic,bhattacharya2024monocular}. Event-based catching has also been transferred to a quadruped~\cite{forrai2023event}. These systems demonstrate rapid sensing, but specialized event sensors are less ubiquitous than commodity RGB cameras, and free-flight or quadrupedal avoidance has different balance constraints from humanoid whole-body evasion.

REBot learns reflexive quadruped evasion, while APREBot adds omnidirectional sensing and active camera focusing~\cite{xu2025rebot,xu2025aprebot}. They share the need to coordinate sensing and evasive motion. Our system studies this coordination on a biped, where search, avoidance, and recovery must all preserve balance with a moving torso camera.

\subsection{Humanoid Ball Interaction}
Humanoid Goalkeeper learns agile reactions from task-motion constraints and motion priors~\cite{ren2025goalkeeper}. Recent badminton, table-tennis, and tennis systems combine trajectory prediction, planning, reinforcement learning, and motion priors to achieve dynamic interaction~\cite{liu2025humanoidbadminton,chen2026badminton,su2025hitter,hu2025pace,ren2026smash,zhang2026latent}. These works primarily intercept or strike the object and commonly rely on external tracking or structured localization. SMASH uses egocentric stereo vision, but also uses fiducials in a constrained table-tennis scene~\cite{ren2026smash}. In contrast, our robot avoids a fast ball across a broad workspace using one onboard monocular RGB camera and no external object tracking.

PAC-MAN directly studies humanoid dodgeball using CBF-guided reinforcement learning, an adversarial motion prior, and segmentation-masked onboard depth~\cite{yang2026pacman}. Its analysis relates useful safety structure to perceptual observability. Our system uses monocular RGB and temporal estimation, and composes learned evasion/recovery with full-azimuth search and tracking. We evaluate the resulting perception--estimation--control pipeline; differing hardware protocols preclude a direct success-rate comparison.

\section{METHODS}
\subsection{System and Task Formulation}
\label{sec:task}
Figure~\ref{fig:overview} summarizes the deployment loop and its training interfaces. The robot is a Unitree G1 with 29 actuated joints and a torso-mounted monocular camera. At time $t$, perception produces a normalized box
\begin{equation}
\label{eq:bbox}
\mathbf b_t=[c^u_t,c^v_t,w_t,h_t,m_t],
\end{equation}
where center and size are normalized by image width and height, and $m_t\in\{0,1\}$ indicates a valid detection. The estimator maps a history $\mathbf b_{t-H+1:t}$ to relative ball position $\hat{\mathbf p}^{r}_{b,t}$ in the robot frame and a predicted binary threat flag $\hat{\chi}_t$. The ground-truth flag $\chi_t=1$ denotes an approaching ball with nonzero closing speed; $\chi_t=0$ denotes a staged, receding, or absent ball. The threat flag indicates whether the ball is approaching rather than whether contact is inevitable. The policy combines estimated task state with proprioception and emits 29 joint-position targets. During training, an asymmetric critic additionally receives privileged ball trajectory, target, phase, time-to-target, and base linear velocity.

\subsubsection{Training environment}
\begin{figure*}[t]
    \centering
    \includegraphics[width=0.9\textwidth]{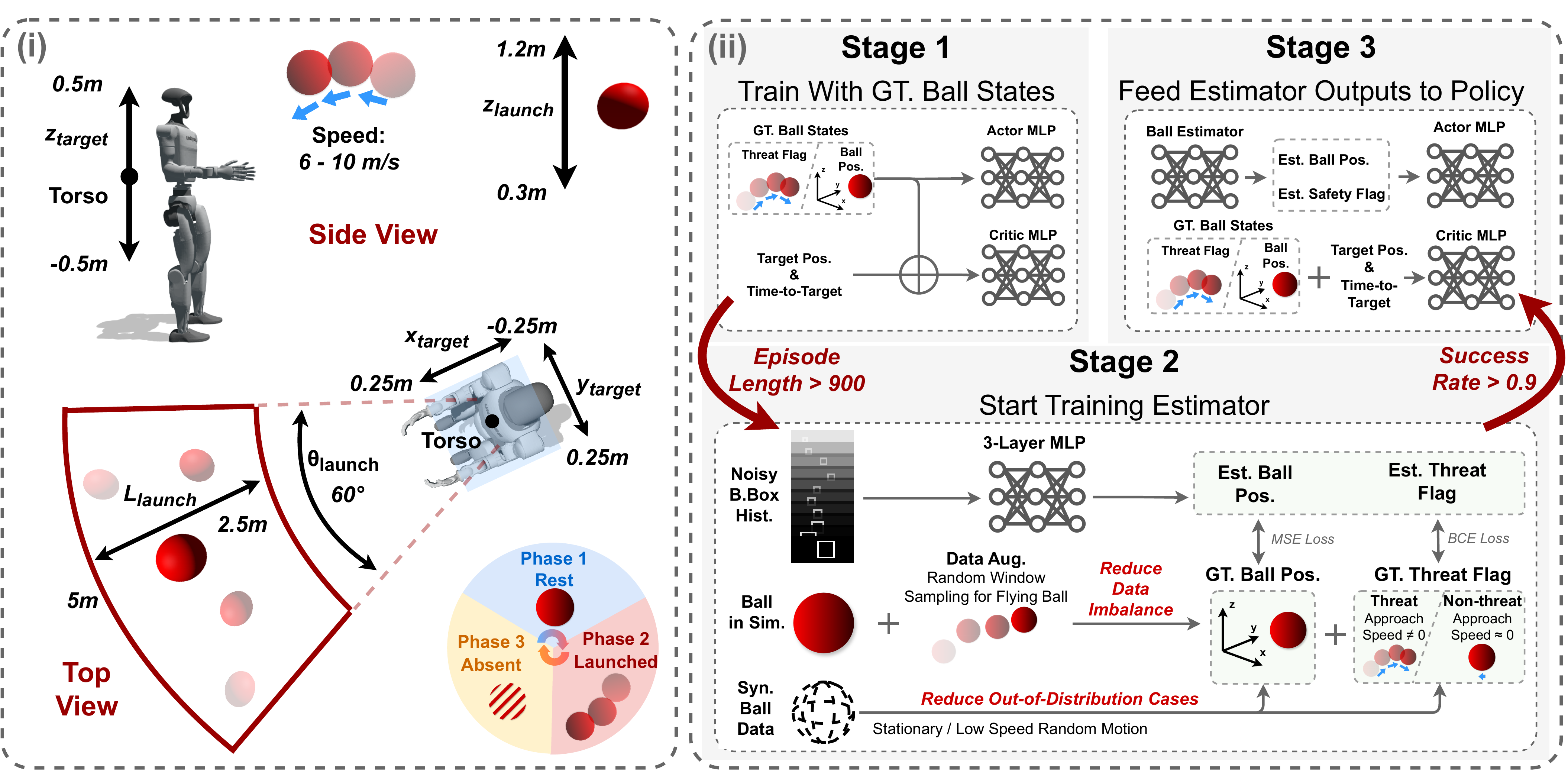}
    \caption{Simulation environment and three-stage curriculum. (i) Each episode samples launch distance, speed, yaw, height, and torso target from the shown ranges, and the ball alternates among rest, launched, and absent phases. (ii) Stage 1 trains the policy with ground-truth (GT) ball position and threat flag. Stage 2 starts supervised estimator training after the episode length exceeds 900 steps while the actor retains these GT inputs. Stage 3 feeds the estimated ball position and threat flag to the actor after dodge success exceeds 0.9; the critic retains privileged training inputs.}
    \label{fig:curriculum}
    \vspace{-1.5em}
\end{figure*}

The training environment presents repeated encounters rather than a single fixed throw. Each episode contains five throws separated by randomized 2--4 s intervals, preventing the policy from exploiting a periodic launch schedule. As shown in Fig.~\ref{fig:curriculum}(i), each ball is initialized in front of the robot within the camera field of view. Launch distance is sampled from 2.5 to 5.0 m, speed from 6 to 10 m/s, yaw over a $60^\circ$ sector, and height from 0.3 to 1.2 m. The target is sampled around the torso with horizontal offsets of up to 0.25 m and a vertical offset in $[-0.5,0.5]$ m. The resulting trajectory diversity discourages convergence to a single fixed evasive motion.

Within each throw, the ball progresses through rest, launched, and absent phases. The rest phase provides visible but non-threatening observations; the launched phase requires the robot to distinguish an approaching ball and evade it; and the absent phase removes the visual target after the encounter. These phases provide both threat and non-threat observations and require the dodge/recovery policy to respond to ball state rather than visibility alone. This cyclic design reflects realistic throwing scenarios, in which the ball is entirely out of view during the reset interval between the robot completing its dodge and the next throw preparing.

\subsection{Monocular Ball Perception}
\label{sec:bbox}
Rendering thousands of camera streams during reinforcement learning is expensive. Instead, we analytically project the known spherical ball in simulation. For ball center $\mathbf p^w_{b,t}$, camera position $\mathbf p^w_{c,t}$, and world-to-camera rotation $\mathbf R_{cw}$,
\begin{equation*}
\mathbf p^c_{b,t}=\mathbf R_{cw}(\mathbf p^w_{b,t}-\mathbf p^w_{c,t})=[x,y,z]^\top.
\end{equation*}
For a sphere fully in front of the camera ($z>r_b$) and either image-plane coordinate $a\in\{x,y\}$ and ball radius $r_b$, the two tangent-ray slopes are
\begin{equation*}
m_a^{\pm}=\frac{az\pm r_b\sqrt{a^2+z^2-r_b^2}}{z^2-r_b^2}.
\end{equation*}
Camera intrinsics map $m_x^{\pm}$ and $m_y^{\pm}$ to pixel extrema. We clip the extrema to the image, convert them to the center-size representation in~\eqref{eq:bbox}, and return a blank box when the sphere is behind the camera or outside its field of view.

At deployment, a fine-tuned YOLOv8 detector~\cite{jocher2023yolov8} supplies the same representation. We collect about two minutes of onboard video, label 830 frames, and expand the training set to 3050 images with cropping, rotation, and illumination augmentation. Detector validation uses 313 images; its measured performance is reported in Section~\ref{sec:perceptioneval}.

Clean analytic boxes do not reproduce hardware errors. During training we therefore (i) randomly drop visible detections, (ii) add Gaussian center noise in pixels, (iii) apply multiplicative log-normal width and height noise, and (iv) delay observations by 30--50 ms. This interface exposes the downstream estimator and policy to missing, shifted, scaled, and stale measurements without rendering RGB images.

\subsection{Temporal Ball-State and Threat-Flag Estimation}
\label{sec:estimator}
A single-step monocular bounding box provides no information about the ball's motion and may be unavailable when the detector misses the ball. We therefore estimate the ball state from a sequence of recent detections using a three-layer multilayer perceptron $f_\phi$:
\begin{equation*}
(\hat{\mathbf p}^{r}_{b,t},\hat{\chi}_t)=f_\phi(\mathbf b_{t-H+1:t}).
\end{equation*}
The center trajectory provides motion cues, while the box size is used to infer depth under the assumption that the ball radius is approximately known. Consequently, deviations from the assumed radius can reduce depth estimation accuracy. Simulation provides relative-position labels $\mathbf p^r_{b,t}$ and binary ground-truth threat flags $\chi_t$. We train with
\begin{equation*}
\mathcal L_{\mathrm{est}}=\lambda_p\|\hat{\mathbf p}^{r}_{b,t}-\mathbf p^{r}_{b,t}\|_2^2
+\lambda_y\mathcal L_{\mathrm{BCE}}(\hat{\chi}_t,\chi_t).
\end{equation*}
The estimator's outputs---relative ball position and a binary threat flag---are used by both control policies. The estimator is not trained to explicitly estimate ball velocity. Instead, the policy can infer the velocity from changes across consecutive estimated ball positions.

Because the ball is in flight for only a small fraction of each episode, non-threat steps substantially outnumber threat steps. To prevent the estimator from learning a trivial non-threat bias, we sample windows around the start, middle, and end of visible flight segments. Window length is randomized from 2 to 20 steps, stride from 1 to 3 steps, and each retained window contains at least two visible frames. We also synthesize non-threatening stationary and low-speed trajectories inside the camera frustum. Depth is log-uniform on 0.2--7.0 m; velocity is either near-static in $[-0.05,0.05]$ m/s with probability 0.6 or sampled in $[-2,2]$ m/s. Normal rollout, flight-window, and synthetic samples are mixed approximately $1:1:2$.

\begin{table}[htbp]
\vspace{-6pt}
\caption{Control objectives and their roles in the two skills.}
\vspace{-4pt}
\label{tab:objectives}
\centering\small
\begin{tabular}{p{0.24\columnwidth}p{0.65\columnwidth}}
\toprule
Component & Objective and intended behavior \\
\midrule
Dodge & Avoid ball contact and increase minimum body-to-ball clearance. \\
Recovery & Reduce planar motion, yaw rate, and deviation from the ready pose after evasion. \\
Search/track & Align horizontal facing direction with the ball to acquire and maintain visibility. \\
Regularization & Discourage unstable or unnecessarily aggressive whole-body motion. \\
\bottomrule
\end{tabular}
\vspace{-12pt}
\end{table}

\subsection{Whole-Body Policy Learning and Skill Composition}
\label{sec:policy}

\subsubsection{Shared observation and action interfaces}
The system uses two separately learned policies, $\pi_{\mathrm{dodge}}$ for dodge/recovery and $\pi_{\mathrm{search}}$ for search/track. Both observe estimated ball-state history, base angular velocity, projected gravity, joint states, and the previous action. Both output joint-position targets for the same 29-joint robot. Each uses an asymmetric actor--critic architecture: the actor is trained toward deployment-available observations, whereas the critic can use privileged task and robot states.

\begin{figure*}[t]
    \centering
    \includegraphics[width=0.9\textwidth]{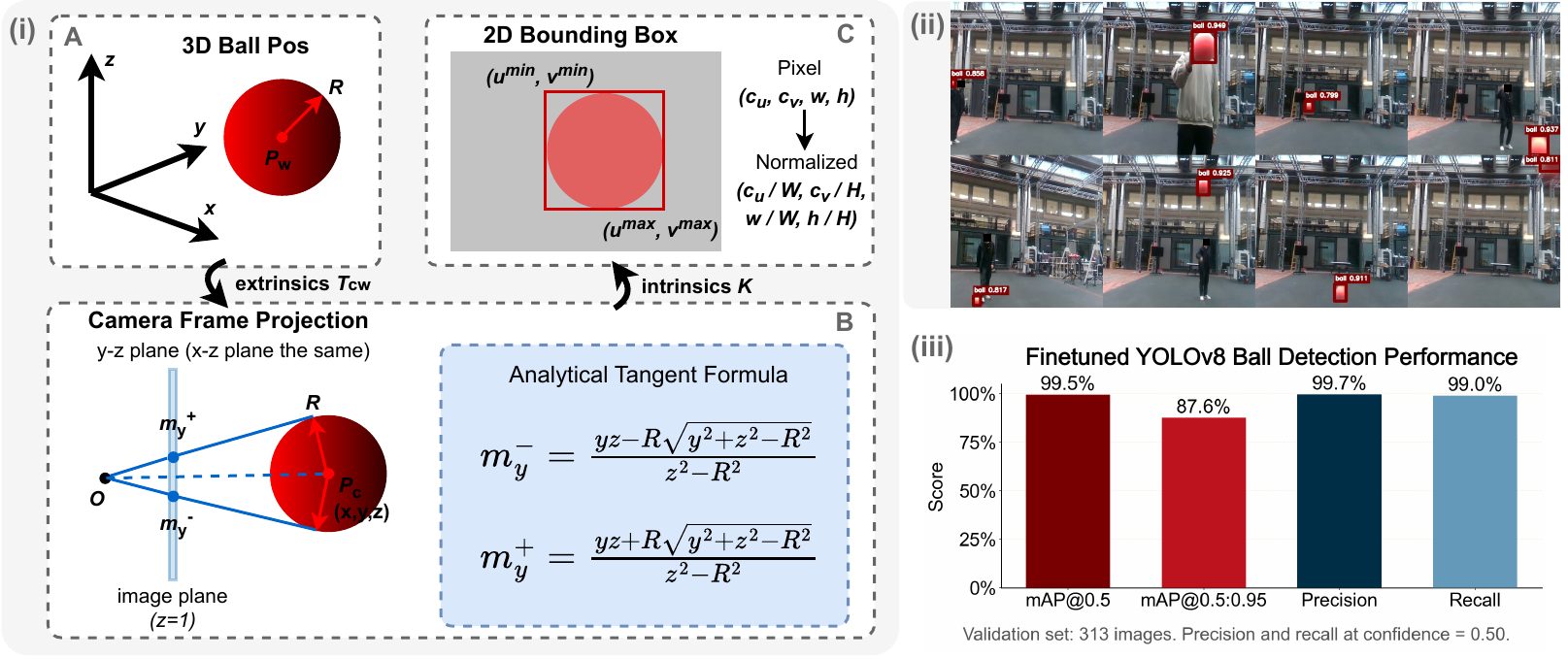}
    \caption{Perception interface. (i) The simulated ball is transformed into the camera frame and projected analytically from tangent rays. (ii) Hardware data are collected from the onboard camera and labeled for a single ball class. (iii) Validation performance of the fine-tuned detector.}
    \label{fig:perception}
    \vspace{-12pt}
\end{figure*}

\subsubsection{Learning dodge and recovery}
The dodge/recovery objective combines collision avoidance and body-to-ball clearance with a return to a stable ready posture after the threat passes. Let $C_t$ indicate contact with the ball, $\mathbf p_{j,t}^{r}$ denote the position of robot body $j$, and $\mathbf q^0$ denote the ready pose. The task terms are
\begin{align*}
r_t^{\mathrm{dodge}}={}&w_{\mathrm{nc}}(1-C_t)-w_{\mathrm c}C_t
+w_{\mathrm d}\min_j\|\mathbf p_{b,t}^{r}-\mathbf p_{j,t}^{r}\|_2,\nonumber\\
r_t^{\mathrm{rec}}={}&-w_v\|\mathbf v_t^{xy}\|_2^2-w_\omega\omega_{z,t}^2
-w_q\sum_j|q_{j,t}-q_j^0|.
\end{align*}
The first line rewards a collision-free response with clearance; the second discourages residual planar motion, yaw, and posture error during recovery. Regularization terms penalize unstable or unnecessarily aggressive actions. Thus, an avoidance that ends in a fall or leaves the robot unable to face the next throw is not rewarded as a complete response. Table~\ref{tab:objectives} summarizes the roles of the policy objectives.

\subsubsection{Learning search and tracking}
The search/track policy aligns the robot's horizontal facing direction with the ball. With unit vectors $\mathbf f_t$ along the robot's horizontal facing direction and $\mathbf d_t$ from the robot to the ball, the principal task reward is
\begin{equation*}
r_t^{\mathrm{face}}=3\exp\!\left[-\frac{(1-\mathbf f_t^\top\mathbf d_t)^2}{0.5^2}\right].
\end{equation*}
The camera is fixed to the torso, so changing its viewing direction requires whole-body turning and stepping. This skill must acquire the target and stabilize the view rather than produce a short avoidance reflex. The two skills remain separate to retain these different temporal behaviors. Hardware observations expose a practical trade-off: faster turning increases angular coverage but can blur the image and interrupt detections.

\subsubsection{Skill composition}
Let $F_t$ denote a hysteretic ball-found state derived from consecutive valid or invalid detections, and let $D_t$ denote the predicted threat flag obtained from $\hat{\chi}_t$. These variables organize the behavior sequence: search/reacquire when the ball is not found, track or recover when it is visible but non-threatening, and evade when it is visible and threatening. The dodge/recovery skill returns the robot toward a ready posture after an avoidance response, while the search/track skill supports subsequent reacquisition. Hysteresis reduces changes in the found state caused by intermittent detections. This is a system-level composition of two policies; it does not imply simultaneous mixing of their joint targets or a single phase-gated policy.

\subsubsection{Three-stage curriculum}
\label{sec:curriculum}
Training the perception-conditioned dodge/recovery policy from the first update would require the robot to learn balance and whole-body evasion while simultaneously relying on an immature estimator. Early falls and unstructured body motion also produce unstable camera trajectories, which are poor supervision for learning the estimator. The curriculum in Fig.~\ref{fig:curriculum}(ii) separates these coupled learning problems and introduces estimated observations only after the underlying control behavior and visual trajectories become sufficiently stable.

\noindent\textbf{Stage 1: ground-truth control warm-up.}
The dodge/recovery actor receives ground-truth relative ball position and threat flag. Removing estimation errors at this stage allows it to first acquire balance-preserving evasion and recovery behavior. The asymmetric critic uses the privileged task and robot states described above.

\noindent\textbf{Stage 2: estimator learning.}
Estimator supervision starts when the episode length exceeds 900 steps, while the actor continues to receive ground-truth ball states. Delaying estimator training until this trigger avoids collecting most of its initial data from short, failure-dominated rollouts. The estimator is trained from the policy-generated bounding-box histories together with the flight-window and synthetic samples described in Section~\ref{sec:estimator}. Keeping ground-truth inputs at the actor also prevents early estimator errors from destabilizing the control policy.

\noindent\textbf{Stage 3: estimator-conditioned control.}
When dodge success exceeds 0.9, the actor's ground-truth ball inputs are replaced by the estimated relative position and threat flag. Estimator supervision remains active as the policy adapts to missed detections, box perturbations, and the 30--50 ms observation delay. The critic retains privileged ball state, target, time-to-target, phase, and base linear velocity, whereas the actor now uses the deployment-available perception interface. This final stage closes the perception--estimation--control loop evaluated in Section~\ref{sec:policyeval}.

\begin{figure*}[!t]
    \centering
    \includegraphics[width=0.9\textwidth]{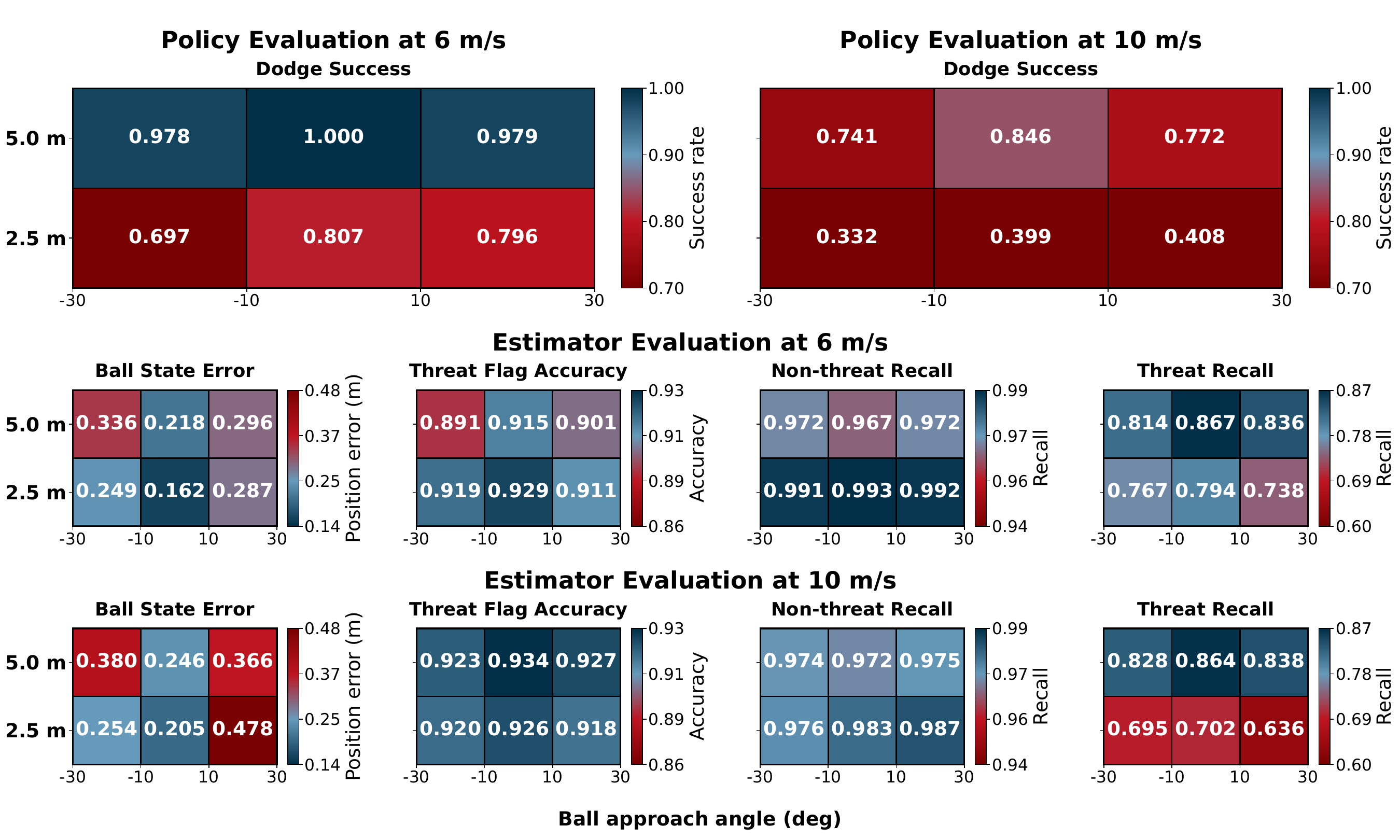}
    \vspace{-6pt}
    \caption{Controlled policy and estimator evaluation at 6 and 10~m/s. Columns denote left, center, and right approach-yaw bins; rows denote launch distances of 5.0 and 2.5~m. Across the displayed conditions, dodge success ranges from 33.2\% to 100.0\%, relative-position error from 0.162 to 0.478~m, and threat recall from 0.636 to 0.867.}
    \label{fig:analysis}
    \vspace{-12pt}
\end{figure*}

\section{EXPERIMENTS}
\subsection{Experimental Setup and Metrics}
We train and evalute policies in Isaac Lab~\cite{mittal2025isaaclab} with Unitree G1. Evaluation uses estimated ball states and proprioception; ground-truth states are used only to compute evaluation metrics. A dodge succeeds only if the ball contacts no body link and the robot does not fall. Search succeeds when an initially invisible ball enters the field of view and a valid detection is maintained. Single-throw success and complete-sequence success are reported separately.

Detector recall is defined as $\mathrm{TP}/(\mathrm{TP}+\mathrm{FN})$, where $\mathrm{TP}$ and $\mathrm{FN}$ denote the numbers of true-positive and false-negative detections, respectively. Recall therefore measures the proportion of ground-truth objects successfully detected by the model. Average precision (AP) summarizes the precision--recall trade-off across detection confidence thresholds and corresponds to the area under the precision--recall curve; with a single object class, mean average precision (mAP) is equivalent to AP. We report $\mathrm{mAP}@0.5$, which evaluates detections using an intersection-over-union (IoU) threshold of 0.5 and thus captures the model's ability to identify objects with reasonably accurate localization, and $\mathrm{mAP}@0.5{:}0.95$, which averages AP over IoU thresholds from 0.5 to 0.95 in increments of 0.05, providing a more stringent measure of both detection and bounding-box localization accuracy.

Controlled simulation tests use speeds of 6 and 10 m/s and distances of 5.0 and 2.5 m. Approach yaw is divided into left $[-30^\circ,-10^\circ]$, center $[-10^\circ,10^\circ]$, and right $[10^\circ,30^\circ]$ bins. Each speed--distance condition uses 200 parallel environments and five throws per environment, giving 1000 throws in total across the three bins. The nominal distance-to-speed ratio ranges from approximately 0.83 s to 0.25 s; it describes the launch time scale rather than measured time to body contact.

\subsection{Perception Accuracy and Efficiency}
\label{sec:perceptioneval}
As shown in Fig.~\ref{fig:perception} (iii), the fine-tuned detector demonstrates reliable ball detection, attaining high scores for all the metrics on the 313-image validation split. Note that these image-level metrics characterize prediction on a static camera, thus decoupling closed-loop measurement during rapid body rotation.

\begin{figure}[!htbp]
    \centering
    \includegraphics[width=1.0\columnwidth]{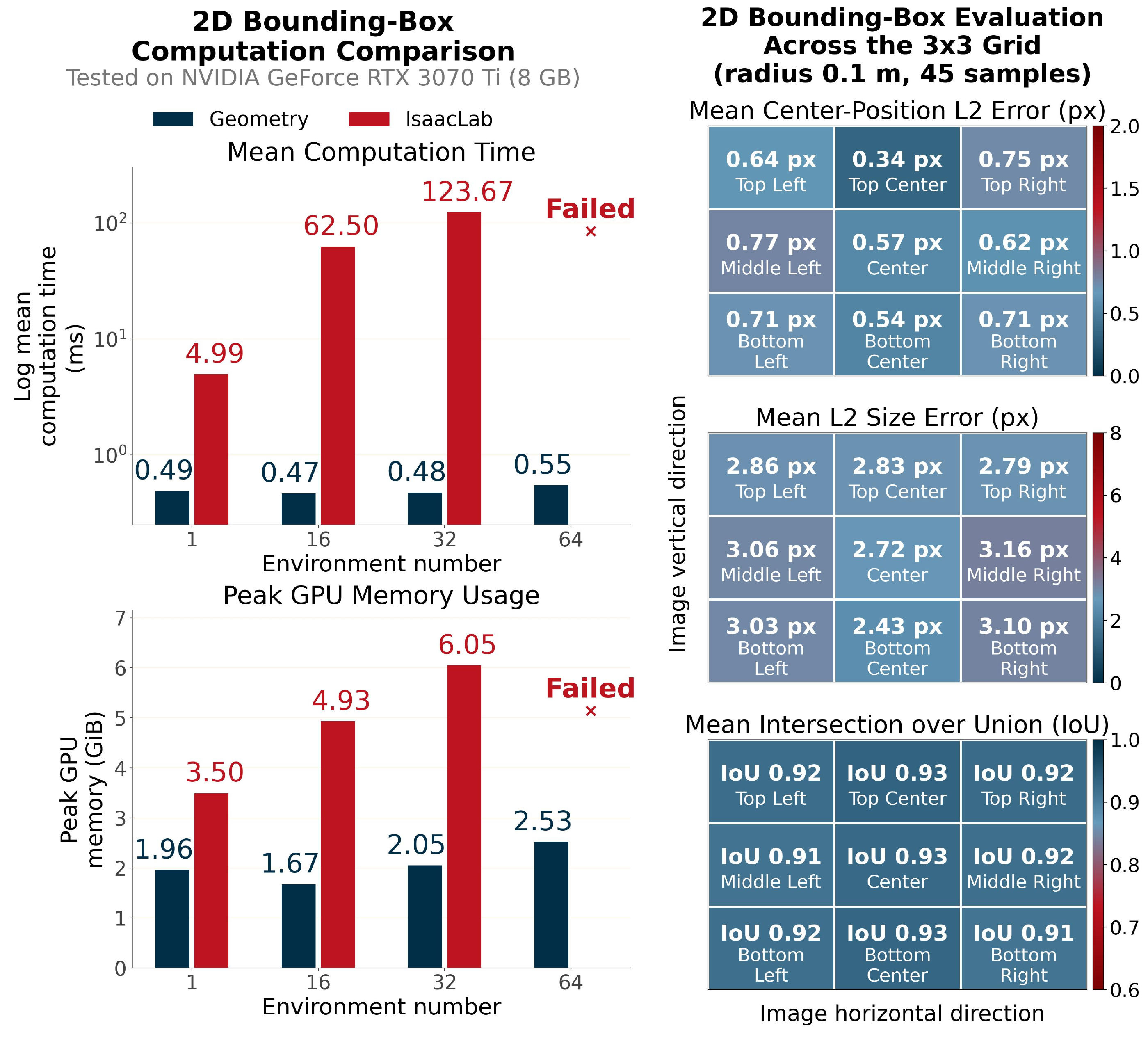}
    \vspace{-6pt}
    \caption{Analytic projection compared with Isaac Lab rendering-based 2D bounding boxes. Left: computation time and peak GPU memory as the number of camera environments increases. Right: center, size, and IoU agreement over a $3\times3$ image grid. The measured speedup applies to observation generation rather than complete policy training.}
    \label{fig:bboxbenchmark}
    \vspace{-16pt}
\end{figure}

\begin{table*}[!htbp]
\vspace{-6pt}
\caption{Robustness to injected 2D bounding-box noise.}
\vspace{-6pt}
\label{tab:bboxnoise}
\centering\small
\begin{tabular}{@{}lccccc@{}}
\toprule
Noise & Dodge (\%) & Error (m) & Threat acc. (\%) & Non-threat rec. (\%) & Threat rec. (\%) \\
\midrule
Clean & 83.53 & 0.1854 & 95.37 & 99.41 & 87.84 \\
Small & 84.40 & 0.2130 & 94.55 & 98.30 & 87.50 \\
Medium & 73.00 & 0.2496 & 91.75 & 94.10 & 87.42 \\
Large & 26.90 & 0.3424 & 87.55 & 89.24 & 84.47 \\
\bottomrule
\end{tabular}
\vspace{-15pt}
\end{table*}

To evaluate the efficiency, we compare analytic projection with Isaac Lab rendering-based 2D bounding-box computation on an NVIDIA RTX 3070 Ti with 8 GB memory (Fig.~\ref{fig:bboxbenchmark}). Analytic projection stays near 0.5 ms from 1 to 64 camera environments. Rendering grows from 4.99 ms at one environment to 123.67 ms at 32 and fails at 64. The measured observation-generation speedup ranges from $10.2\times$ to $259.7\times$, with a 1.8--2.9-fold reduction in peak memory. These measurements concern bounding-box generation, not total reinforcement-learning training time.

Across a $3\times3$ image grid, the mean center error is below one pixel, the mean size error is approximately 2.4--3.2 pixels, and the mean IoU is 0.912--0.933. Thus, the analytic interface closely approximates the tested rendered boxes while avoiding image rendering during policy training.

\subsection{Ball-State and Threat-Flag Estimation}
\label{sec:esteval}
We denote ball position estimation error in robot frame by $\|\hat{\mathbf p}^{r}_{b,t}-\mathbf p^{r}_{b,t}\|_2$. Threat-flag accuracy is defined as the fraction of correct binary predictions. Threat recall is recall for $\chi_t=1$, whereas non-threat recall is recall for $\chi_t=0$.

Across the controlled estimator evaluation tests in Fig.~\ref{fig:analysis} (lower two), the mean position error ranges from 0.162 to 0.478~m and threat-flag accuracy from 0.891 to 0.934. At 6~m/s, reducing the launch distance from 5.0 to 2.5~m lowers threat recall; at 10~m/s, it falls further. Non-threat recall remains between 0.967 and 0.993. The 2.5~m, 10~m/s condition therefore has the lowest threat recall and the widest position-error range, identifying high difficulty at setups with high ball speed and short distance.

To evaluate the ball-state estimator performance in both simulation and real-world conditions, we use the iteration-10k simulation policy with injected bounding box noise and calibrate the real-world noise level on a test held-out dataset. While the clean setting applies no perturbation, small, medium, and large noise use pixel-center standard deviations of 2, 16, and 32 pixels, relative-center standard deviations of 0.05, 0.40, and 0.80, and log-size standard deviations of 0.10, 0.80, and 1.60, respectively. Each setting aggregates 3 seeds and 3000 throws. Table~\ref{tab:bboxnoise} reports success rate, position error, and threat-flag metrics.

As the noise grows, the estimation position error increases monotonically, while threat accuracy decreases. Small noise has little effect on dodge success, whereas medium and large noise reduce it to 73.00\% and 26.90\%, respectively. Calibration of YOLO outputs on the validation set shows that real-image bounding-box noise is close to the small setting and falls below it when the scene and illumination match the training distribution. During deployment, environmental shifts can increase the observed noise to between the small and medium settings. Policy success in this regime is close to the hardware result, suggesting that this range captures the relevant real-world operating conditions. The policy therefore tolerates nominal perturbations but is sensitive to larger box corruption. Since throws are randomized rather than paired, these results describe distribution-level robustness without establishing pairwise statistical significance.

\subsection{Policy Curriculum Ablation}
\label{sec:policyeval}

We ablate the transitions during policy training detailed in Section~\ref{sec:curriculum} and show results in Table~\ref{tab:policyablation}.
\textit{Full Stages} is the complete curriculum. \textit{Direct Training} uses supervision, perturbations, and estimated actor inputs from the first update. \textit{No Bootstrap} trains the estimator from the outset but retains ground-truth actor inputs until the success threshold. \textit{No Estimator Warmup} activates estimator training, perturbations, and estimated actor inputs together at the episode-length threshold.
The evaluation fixes the target at the torso origin, samples speed from $[6,10]$~m/s, and retains nominal randomization. 
Each result aggregates 1000 throws and reports strict success and mean position error over valid frames.

The full curriculum attains 90.8\% success, demonstrating a strong performance above the ablations. \textit{No Bootstrap} has the lowest position error but lower success. \textit{No Estimator Warmup} has both the lowest success and the largest error.

\begin{table}[!htbp]
\vspace{-7pt}
\caption{Curriculum ablation.}
\vspace{-6pt}
\label{tab:policyablation}
\centering\small
\begin{tabular}{@{}lcc@{}}
\toprule
Training variant & Dodge (\%) & Error (m) \\
\midrule
Full Stages & 90.8 & 0.2614 \\
Direct Training & 74.1 & 0.2742 \\
No Bootstrap & 80.6 & 0.2152 \\
No Estimator Warmup & 67.6 & 0.5572 \\
\bottomrule
\end{tabular}
\vspace{-12pt}
\end{table}

\subsection{Integrated System and Hardware Evaluation}
\subsubsection{Controlled simulation performance}
Figure~\ref{fig:analysis} shows a clear dependence of policy performance on the available response time. At 6~m/s, reducing launch distance from 5.0 to 2.5~m lowers success across approach-angle bins. At 10~m/s, success decreases further. The hardest condition has a nominal distance-to-speed ratio of 0.25~s, together with the lowest threat recall and the widest position-error range. The degradation therefore reflects a combined sensing-and-control challenge rather than a single estimator metric.

The dodge/recovery policy avoids ten consecutive throws in a simulation sequence. Separately, the search/track policy finds balls initialized across $[-180^\circ,180^\circ]$ in ten of ten resets. These are small behavior demonstrations, distinct from the 1000-throw controlled evaluations. Table~\ref{tab:endtoend} summarizes these simulation demonstrations alongside the available real-robot evaluation results.

\subsubsection{Real-robot deployment}
We deploy the trained policy to a real Unitree G1 robot and refer the hardware results to our supplementary videos. The physical G1 uses no external object tracking: the RGB detector supplies box histories, and the active policy executes joint targets at 50 Hz. This rate does not measure detector throughput or end-to-end latency. Strict no-contact/no-fall success exceeds 85\% over more than 15 controlled single throws; exact counts and conditions will replace this aggregate after record consolidation.

A hardware sequence in Fig.~\ref{fig:hardware} demonstrates track--dodge--search--dodge: The robot first tracks the ball held by a moving thrower to maintain visual observation and evades it once it is thrown; It then performs a full-azimuth search, reacquires the ball outside the camera view, stabilizes tracking, and dodges again. Policies use only onboard monocular RGB-derived state and proprioception.
In our test, we find search behavior remains sensitive to motion blur. Rapid torso rotation expands angular coverage but produces intermittent detections; slower turning improves visibility while delaying reacquisition. These observations link search motion quality to the sensing conditions required for the next dodge.

\begin{table}[!htbp]
\vspace{-6pt}
\caption{System evaluation.}
\vspace{-6pt}
\label{tab:endtoend}
\centering\small
\begin{tabular}{@{}llc@{}}
\toprule
Domain & Task & Result \\
\midrule
Simulation & Consecutive dodge & 10/10 \\
Simulation & Full-azimuth search & 10/10 \\
Hardware & Single throws & $>85\%$, $N>15$ \\
Hardware & Reacquisition + dodge & Suppl. Video \\
\bottomrule
\end{tabular}
\vspace{-12pt}
\end{table}

\subsection{Limitations}
Despite the effectiveness of the perception pipeline in our experiments, several limitations remain. The estimator assumes that the ball's approximate physical size are known in advance. This prior knowledge provides a useful monocular depth cue, but estimation accuracy may decrease when the target differs from the assumed size.
Although the hardware experiments demonstrate the feasibility of the integrated system, experiments involving consecutive throws are needed to evaluate its robustness across repeated perception, dodge, recovery and search cycles. Future work could provide a more precise assessment by comparing its ball-position estimates with ground-truth measurements obtained from a motion-capture system. Finally, the learned policy often favors a single dodge direction, even when other effective motions may be available. Future work could train complementary dodge policies and distill them into a unified policy, enabling a broader range of dodge motions and more adaptive responses to different incoming trajectories.

\section{CONCLUSION}
This work presents an integrated humanoid dodgeball system combining onboard monocular perception, temporal ball-state and threat estimation, and learned whole-body control. To support efficient large-scale training, bounding-box observations are generated through analytic geometric projection rather than image rendering, enabling parallel simulation while preserving the visual information required by the temporal estimator. Using a history of monocular bounding boxes, the estimator infers the ball position and determines whether it poses an immediate threat. A three-stage curriculum then gradually replaces privileged ball states with estimated states, allowing the dodge/recovery policy to adapt to perception errors before deployment. A separate search/track policy further enables the robot to reacquire and track balls outside its current field of view. Together, these components form an extended behavior loop comprising ball detection, dodging, posture recovery, and target reacquisition. Experiments in simulation and on a physical Unitree G1 demonstrate the effectiveness of the designed system. In simulation, dodge success ranges from 97.8--100.0\% across approach-angle bins for 5.0~m launches at 6~m/s. On the physical robot, in controlled single-throw trials, the system achieves a no-contact/no-fall success rate above 85\% over repeated throws. The hardware experiments also include sequences in which the robot reacquires the ball and successfully responds to a subsequent throw. While close-range and high-speed launches remain challenging, the results demonstrate the feasibility of an onboard sensing-to-action loop that integrates ball estimation, whole-body avoidance, recovery, and target reacquisition, providing a foundation for humanoid robots to respond robustly to fast-moving objects under realistic sensing and balance constraints.

\section*{ACKNOWLEDGMENTS}
{\small
The authors thank Junzhe He, Zhihao Cao, and Hanyu Wu at ETH Zurich for their valuable suggestions and assistance with the hardware experiments. The authors used generative AI to assist with drafting and editing the manuscript. All content was reviewed by the authors, who take full responsibility for it.
}

\bibliographystyle{ieeetr}
\bibliography{references}

\end{document}